\documentclass{article}
\usepackage{amsmath,graphicx,hyperref}
\usepackage[utf8]{inputenc}
\usepackage[T1]{fontenc}
\usepackage{booktabs}
\usepackage{multirow}
\usepackage{multicol}

\usepackage{xcolor}
\usepackage{amssymb}
\usepackage{graphicx}

\usepackage{multirow}
\usepackage{makecell}
\usepackage{xcolor}
\usepackage{setspace}

\newcommand{\NoteSmall}{\fontsize{7.8pt}{9.1pt}\selectfont}

\definecolor{cb_blue}{HTML}{4477AA}
\definecolor{cb_green}{HTML}{117733}
\definecolor{cb_pink}{HTML}{CC6677}
\definecolor{cb_yellow}{HTML}{CCBB44}
\definecolor{cb_orange}{HTML}{EE7733}
\definecolor{cb_teal}{HTML}{44AA99}
\definecolor{cb_purple}{HTML}{882255}

\definecolor{image1_color_green}{HTML}{4EA72E}
\definecolor{image1_color_red}{HTML}{C00000}

\usepackage{tikz}
\usetikzlibrary{shapes.geometric}

\DeclareRobustCommand{\LegendCaptionLine}[1]{
    \tikz[baseline=-0.6ex]{
        \draw[#1, line width=0.8pt] (-1.8ex, 0ex) -- (1.8ex, 0ex);
    }}

\DeclareRobustCommand{\LegendCaptionCircle}[1]{
    \tikz[baseline=-0.6ex]{
        \node[circle, fill=#1, draw=#1, inner sep=0pt, minimum size=1.2ex] at (0,0ex) {};
    }}

\DeclareRobustCommand{\LegendCaptionSquare}[1]{
    \tikz[baseline=-0.6ex]{
        \node[rectangle, fill=#1, draw=#1, inner sep=0pt, minimum width=1.1ex, minimum height=1.1ex] at (0,0ex) {};
    }}

\DeclareRobustCommand{\LegendCaptionTriangle}[1]{
    \tikz[baseline=-0.6ex]{
        \node[regular polygon, regular polygon sides=3, rotate=0, fill=#1, draw=#1, inner sep=0pt, minimum size=1.7ex] at (0,-0.1ex) {};
    }}

\DeclareRobustCommand{\LegendCaptionLineStar}[2]{
    \tikz[baseline=-0.6ex]{
        \draw[#2, line width=0.8pt] (-1.8ex, 0ex) -- (1.8ex, 0ex);
        \node[star, star points=5, star point ratio=2.25, fill=#1, draw=#1, inner sep=0pt, minimum size=1.7ex] at (0, -0.1ex) {};
    }}

\DeclareRobustCommand{\LegendCaptionLineDiamond}[2]{
    \tikz[baseline=-0.6ex]{
        \draw[#2, line width=0.8pt] (-1.8ex, 0ex) -- (1.8ex, 0ex);
        \node[diamond, fill=#1, draw=#1, inner sep=0pt, minimum width=1.2ex, minimum height=1.9ex] at (0,0ex) {};
    }}

\DeclareRobustCommand{\LegendCaptionLineTriangleDown}[2]{
    \tikz[baseline=-0.6ex]{
        \draw[#2, line width=0.8pt] (-1.8ex, 0ex) -- (1.8ex, 0ex);
        \node[regular polygon, regular polygon sides=3, rotate=180, fill=#1, draw=#1, inner sep=0pt, minimum size=1.9ex] at (0,0.1ex) {};
    }}

\DeclareRobustCommand{\LegendCaptionLinePentagon}[2]{
    \tikz[baseline=-0.6ex]{
        \draw[#2, line width=0.8pt] (-1.8ex, 0ex) -- (1.8ex, 0ex);
        \node[regular polygon, regular polygon sides=5, fill=#1, draw=#1, inner sep=0pt, minimum size=1.7ex] at (0,-0.1ex) {};
    }}

\usepackage[preprint]{spconf}
\toappear{
\hfill
\begin{minipage}{0.48\textwidth} 
\vspace{-0cm}
\tiny
\copyright\ 2026 IEEE. Personal use of this material is permitted. Permission from IEEE must be obtained for all other uses, in any current or future media, including reprinting/republishing this material for advertising or promotional purposes, creating new collective works, for resale or redistribution to servers or lists, or reuse of any copyrighted component of this work in other works.
\end{minipage}
}
\ninept 

\title{Which private attributes do VLMs agree on and predict well?}

\name{Olena Hrynenko$^{1,2}$, Darya Baranouskaya$^{1,2}$, Alina Elena Baia$^{2}$, Andrea Cavallaro$^{1,2}$}
\address{$^{1}$EPFL, Switzerland, $^{2}$Idiap Research Institute, Switzerland}

\begin{document}
\pagenumbering{gobble}

\maketitle
\vspace{-1pt}
\begin{abstract}
Visual Language Models (VLMs) are often used for zero-shot detection of visual attributes in the image. 
We present a zero-shot evaluation of open-source VLMs for privacy-related attribute recognition. 
We identify the attributes for which VLMs exhibit strong inter-annotator agreement, and discuss the disagreement cases of human and VLM annotations. 
Our results show that when evaluated against human annotations, VLMs tend to predict the presence of privacy attributes more often than human annotators. In addition to this, we find that in cases of high inter-annotator agreement between VLMs, they can complement human annotation by identifying attributes overlooked by human annotators.
This highlights the potential of VLMs to support privacy annotations in large-scale image datasets.

\end{abstract}

\vspace{-2pt}

\begin{keywords}
VLMs, attributes recognition, privacy
\end{keywords}
\vspace{-5pt}

\section{Introduction}
\label{sec:intro}
\begin{figure}
    \centering
    \includegraphics[width=0.70\linewidth]{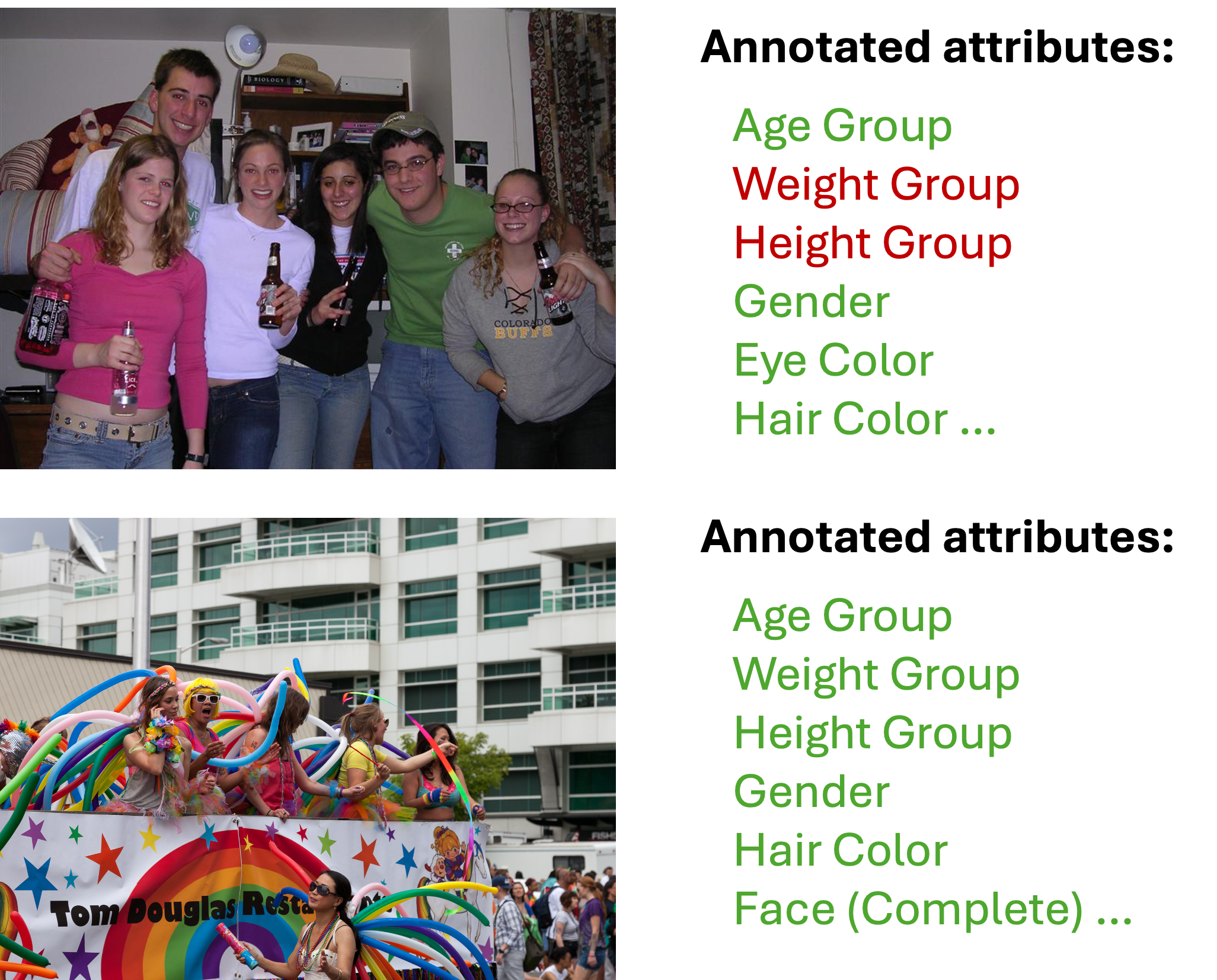}
    \vspace{-10pt}
    \caption{Examples of images from the VISPR dataset~\cite{orekondy_towards_2017}. Attributes annotated as \textcolor{image1_color_green}{present} are shown in \textcolor{image1_color_green}{green}, and those annotated as \textcolor{image1_color_red}{absent} are shown in \textcolor{image1_color_red}{red}. While both images show a group of people, the \textit{Weight Group} and \textit{Height Group} attributes have been omitted by a human annotator for the image in the first row. }
    \label{fig:vispr_missed_examples}
    \vspace{-14pt}
\end{figure}

\vspace{-4pt}
Human labels are the reference for privacy attributes in images. Multi-label annotation of a large number of images (e.g.,~$103$ attributes and approximately $7,000$ images per annotator\footnote{In the~VISPR~dataset~\cite{orekondy_towards_2017}, three annotators performed an image multi-labelling for a dataset of 22,167 images from Flickr, one annotator per image. 
} in the VISPR dataset~\cite{orekondy_towards_2017}) causes a high cognitive load and leads to annotator fatigue~\cite{kost_impact_2018}. On average, a person recalls up to seven concepts in their active memory~\cite{miller_magical_1956}, hence there is a high chance that while the annotated image attributes, or concepts, are correct, they are not the only attributes that are present (see Fig.~\ref{fig:vispr_missed_examples}), leading to false~negatives. 

Recent research has demonstrated that Large Language Models (LLMs) and Visual Language Models (VLMs) are emerging as viable substitutes for human annotators~\cite{chiang-lee-2023-llms_annotators, lu2024visionlanguagemodelsreplacehuman}. For example, using VLMs in image annotation tasks, such as those conducted on the CelebA~\cite{celebA} dataset, shows that automated annotation outputs can achieve high agreement with humans on objective binary classification tasks (e.g., "Is this person wearing eyeglasses? Answer with only yes or no."), while significantly reducing cost~\cite{lu2024visionlanguagemodelsreplacehuman}.
Motivated by the limitations of human annotation and the capabilities of VLMs, in this work, we investigate whether VLMs can recognise a large set of privacy attributes and how their predictions align with human annotations.
Our study focuses on the privacy attributes defined in the VISPR dataset~\cite{orekondy_towards_2017}, which were derived from regulations, social network policies, and the personal judgment of the authors~\cite{orekondy_towards_2017}. 
Examples of attributes include \textit{Gender}, \textit{Signature}, \textit{Full Name}, 
and \textit{Tattoo}.

\vspace{-9pt}
Similar to Chiang and Lee~\cite{chiang-lee-2023-llms_annotators}, to ensure that the experimental setup is similar for both human and VLM annotators, we replicate the annotation instructions provided to the annotators when prompting VLMs.  However, we adapt the setup to a multimodal context. Building on prior work that assesses the ability of VLMs to detect or extract private information in images~\cite{zhang_multi-p2_2024, MultiTrust, REVAL, pixels_vs_privacy, VIP}, we extend the evaluation to the more fine-grained task of private attribute recognition.
Unlike related works~\cite{zhang_multi-p2_2024, MultiTrust, REVAL, pixels_vs_privacy}, we use the entire VISPR~\cite{orekondy_towards_2017} image test set for the evaluation of VLMs' zero-shot performance, consider the full list of privacy attributes provided by VISPR, and use the annotation instructions provided in the VISPR dataset to prompt the models.
Only a subset of attributes
or images is usually considered~\cite{zhang_multi-p2_2024, MultiTrust, pixels_vs_privacy}.
In addition, differently from prior work, we also consider potential human errors in the annotation of the VISPR dataset. 
While~\cite{lu2024visionlanguagemodelsreplacehuman} 
evaluated personal appearance attributes of celebrities, we consider attributes and images that describe not only humans, but also the objects, or locations, widening the set of possible attributes.

\vspace{-1pt}
In summary, we evaluate the ability of VLMs to recognise private attributes and analyse annotation disagreements between the VLMs and human annotators in the VISPR dataset.

\vspace{-1pt}
%%%%%%%%%%%%%%%%%%%%%%%%%%%%%
\section{Related Work}
\label{sec:related_work}
\vspace{-5pt}

Privacy-related studies examine the inference-time privacy risks of VLMs, such as identifying the presence~\cite{ zhang_multi-p2_2024, MultiTrust} or extracting private information ~\cite{MultiTrust, REVAL, pixels_vs_privacy, VIP}.

\vspace{-1pt}
\begin{spacing}{1.0}
The Multi-$P^2A$ benchmark~\cite{zhang_multi-p2_2024} includes measuring VLMs' ability to recognise the presence of any private attribute in the image. 
For this, Zhang~et~al.~\cite{zhang_multi-p2_2024} build a balanced dataset with an equal number of images that contain and do not contain privacy-sensitive content. 
They select 23 from the original 67 VISPR~\cite{orekondy_towards_2017} privacy attributes, excluding some categories on which VLMs can fail, such as \textit{Eye Color}. 
The authors~\cite{zhang_multi-p2_2024} evaluate whether at least one privacy attribute is present, rather than differentiating among specific privacy attributes. 
REVAL~\cite{REVAL} tests VLMs on various personally identifiable information (PII), such as \textit{Home Address}, \textit{Credit Card Number}, \textit{Telephone Number}, and observes that these models struggle to accurately recognise specific privacy attributes in images. Based on this, Zhang~et~al.~\cite{REVAL} propose to reformulate the evaluation as a binary task that assesses only the presence or absence of any privacy attribute.
Similarly, MultiTrust~\cite{MultiTrust} prompts VLMs to make a binary decision about the presence of private information, using a small set of images derived from VISPR~\cite{orekondy_towards_2017} and  VizWiz-Priv~\cite{Gurari_2019_CVPR} covering a limited range of PII categories.

Tömekçe~et~al.~\cite{VIP} show that private attribute inference from social media images is already feasible and that safeguards can be bypassed with simple prompt engineering. Schultenkämper and Bäumer~\cite{pixels_vs_privacy} construct a vision-question answering dataset with prompts crafted for extracting private information. The authors~\cite{pixels_vs_privacy} then assess the ability of VLMs to extract such information, focusing only on directly visible personal attributes, while excluding attributes like \textit{Full Name} or \textit{Place of Birth}. 
MultiTrust~\cite{MultiTrust} evaluates the ability of VLMs to extract private information considering a limited set of PIIs (such as \textit{Email}, \textit{Name}, \textit{Address}, \textit{Credit Card}, and \textit{ID Number}) and a small set of images derived from VISPR. In addition, MMDT~\cite{MMDT} and GeoLocator~\cite{geolocator}  highlight that VLMs can accurately predict locations from landmarks and street cues, exposing geographic privacy.
RTVLM~\cite{rtvlm} shows that many open-source VLMs fail to refuse privacy-sensitive queries, often disclosing personal or celebrity data.

While several studies~\cite{zhang_multi-p2_2024, MultiTrust, REVAL,pixels_vs_privacy, MMDT} reuse VISPR attributes to construct privacy-sensitive probes and test VLMs, they focus on a limited number of private attributes/PIIs and do not evaluate fine-grained recognition of specific privacy attributes.
Our work addresses this limitation by systematically evaluating VLMs on all VISPR privacy attributes using the instructions as for human annotators,   and by analysing the discrepancies between VLMs and human annotators.

\end{spacing} 

\vspace{-9pt}
\section{Experimental setup}
\label{sec:methodology}

\begin{spacing}{1.0}
We use three open-source instruction-following VLMs, namely, Gemma-3-4b-it~\cite{kamath2025gemma}, 
Qwen2.5-VL-7B-Instruct~\cite{Qwen2.5-VL}, 
and Llama-3.2-11B-Vision-Instruct~\cite{grattafiori2024llama}.
The experiments are run on a single RTX3090. For each image in the VISPR test dataset (8000 images), we run a prompt for each of the 67 privacy attributes in a zero-shot setup. The prompt is formulated as follows:
\end{spacing}

\noindent\colorbox{yellow!15}
{
\begin{minipage}{0.95\linewidth}
\ttfamily
\NoteSmall
\textbf{User prompt: }

Here is information about the attribute:

Attribute: \{attribute\}

Attribute's definition: \{description\}
\newline

By using this definition for any subject in the image – either in the foreground or background, answer the following question: is {attribute} present or absent in the image? 

Please answer [Present] or [Absent].

\end{minipage}
}

\begin{spacing}{1.027}
Since the response to the prompt varies from a single word (e.g.,~"Present") to one or more sentences (e.g.,~"The image shows two nail polish designs, but there is no signature present in the image."), we use a two-step process to parse the response into a label. As in~\cite{li_privaci-bench_2025}, we use the same model for parsing the response. We first use the following prompt:
\end{spacing}

\noindent\colorbox{yellow!15}
{
\begin{minipage}{0.95\linewidth}
\ttfamily
\NoteSmall
\textbf{System prompt: }\\
You are provided with an open-ended reply. The possible closed-ended answers are: [present], [absent]. Map open-ended reply to the closed-ended answer. You can reply with: [present], [absent] or [NONE] if none of the closed-ended replies is suitable. You MUST begin your reply with: 'THE ANSWER: ' 
\\

\textbf{User Prompt:}

Open-ended reply: {\{reply\}}
\end{minipage}
}

\vspace{5pt}

 We then apply logic-based substring parsing to obtain the present, absent, or NONE labels. We included the NONE case to account for the inconclusive model's responses (e.g.,~"The image shows a ticket for a Metal Hammer Radio event in Barcelona on June 6, 2002. ... However, there is no indication of Sexual Orientation present or absent in the image."). We enable stochastic decoding (set \texttt{do\_sample=True} in the model's generation parameters), which allows the model to sample from the output distribution and generate more diverse predictions, thereby reducing the risk of consistently predicting the same class when multiple outcomes are considered plausible.

\vspace{-11pt}
\section{Evaluation}
\label{sec:evaluation}
\vspace{-1.5pt}
\subsection{VLMs recognition of private attributes}
\begin{figure}[t!]
    \centering
    \includegraphics[width=0.48\textwidth]{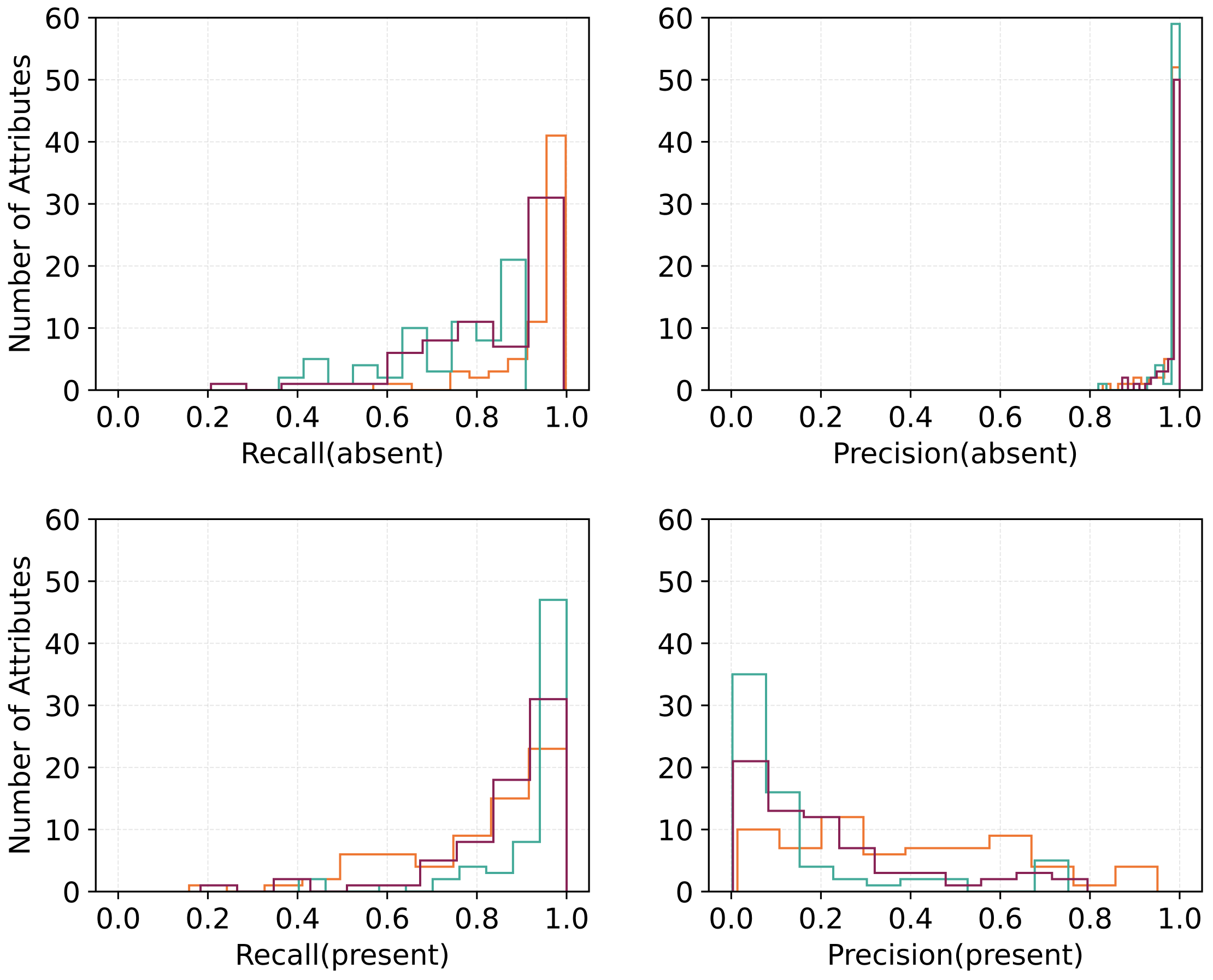}
    \vspace{-19pt}
    \caption{Distribution of precision and recall for present and absent labels for zero-shot recognition of Qwen2.5-VL-7B-Instruct (\hspace{-0.6ex}\protect\LegendCaptionLine{cb_orange}), Gemma-3-4b-it (\hspace{-0.6ex}\protect\LegendCaptionLine{cb_teal}), Llama-3.2-11B-Vision-Instruct (\hspace{-0.6ex}\protect\LegendCaptionLine{cb_purple}) for 67 attributes of the VISPR test set~\cite{orekondy_towards_2017}.
    }
    \label{fig:distributions}
    \vspace{-15pt}
\end{figure}

\begin{figure}[th!]
    \centering
    \includegraphics[width=0.87\linewidth]{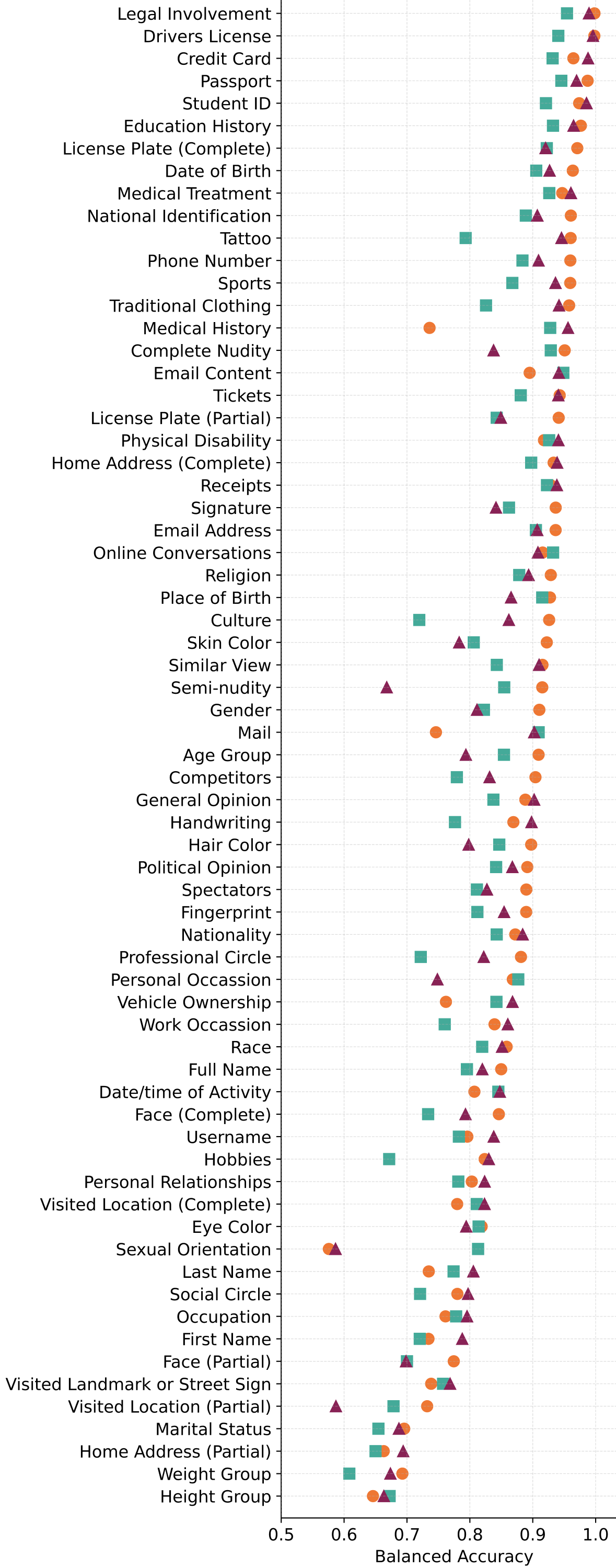}
    \vspace{-10pt}
    \caption{Balanced accuracy of zero-shot privacy attribute recognition for Qwen2.5-VL-7B-Instruct (\hspace{-0.85ex}\protect\LegendCaptionCircle{cb_orange}),
    Gemma-3-4b-it (\hspace{-0.8ex}\protect\LegendCaptionSquare{cb_teal}), and Llama-3.2-11B-Vision-Instruct (\hspace{-0.65ex}\protect\LegendCaptionTriangle{cb_purple}) on the VISPR test set~\cite{orekondy_towards_2017}. }
    \vspace{-17pt}
\label{fig:balanced_accuracy_models}
\end{figure}

\vspace{-1.5pt}
We analyse the distributions of the precision and recall scores for present and absent classes (see Fig.~\ref{fig:distributions}). Additionally, we evaluate how well VLMs recognise privacy attributes by considering the balanced accuracy (see Fig.~\ref{fig:balanced_accuracy_models}).


Models have high recall for both present and absent labels, indicating that they rarely miss attributes when they are actually present in the image and correctly identify truly absent attributes. 
While the precision for the absent class is consistently close to 1, the precision for the present class is notably lower, with models often detecting attributes that are not annotated in the dataset. This discrepancy may not solely reflect model error, given the high cognitive load of labelling a large set of attributes for human annotators. 
As a result, for the majority of attributes and models, the balanced accuracy is above 0.75. However, low precision negatively affects the F1-macro scores, which, for the worst-performing model Gemma-3-4b-it, range from 0.4 to 0.6 across the majority of attributes (57 out of 67).
Across the three models, Qwen2.5-VL-7B-Instruct significantly outperforms Gemma-3-4b-it and Llama-3.2-11B-Vision-Instruct, reaching the highest F1-macro score for the majority of attributes (64 out of 67) and F1-macro over 0.6 for 55 attributes. Overall, despite some limitations in precision for the present class, the performance of Qwen2.5-VL-7B-Instruct aligns the most with annotators' labels, while Gemma-3-4b-it and Llama-3.2-11B-Vision-Instruct perform noticeably worse.

\begin{figure*}
    \centering
    \includegraphics[height=3.4cm]{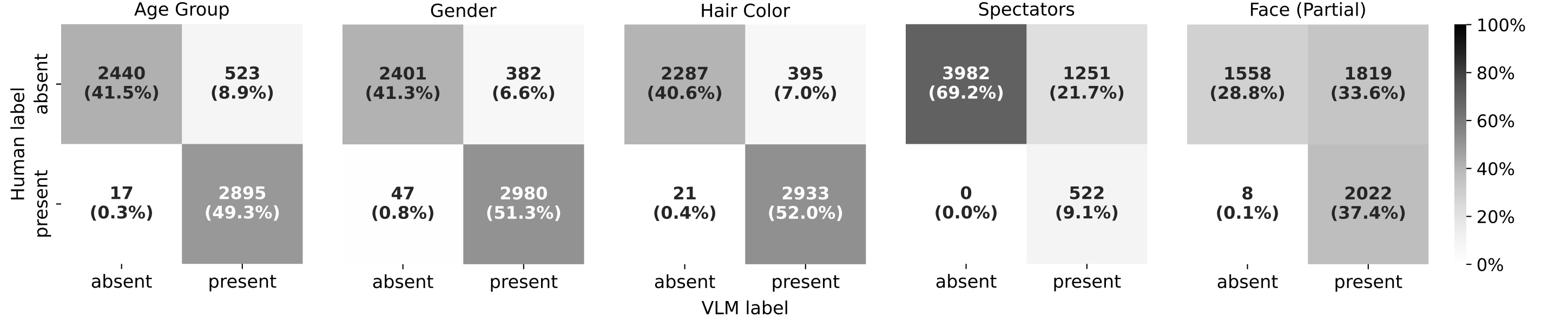}
    \vspace{-12pt}
    \caption{The disagreements between the human labels and VLM labels. We note that there are numerous disagreements when the VLMs predict the presence of an attribute, when, according to the human annotators, the attribute is absent. Integers denote image counts, with proportions shown in parentheses.}
    \label{fig:confusion_matrix}
\vspace{-7pt}
\end{figure*}
\vspace{-8pt}
\subsection{VLMs and human disagreements}
\vspace{-3pt}
\begin{table}[t!]
    \caption{Top-five attributes with the highest VLM annotation agreement according to Fleiss kappa (1 -- perfect agreement, 0 -- annotation agreement by chance). For each attribute, we detail the cases in which VLMs identify it as present, whereas human annotators as absent. We manually inspect 50 images per attribute, report the percentage of cases in which VLMs are correct, and list the image categories in which VLMs are incorrect, from most common to least.}
    \vspace{3pt}
    \centering
    \begin{tabular}{p{1.45cm}p{0.6cm}p{0.96cm}p{3.9cm}}
    \toprule
     &  & \multicolumn{2}{c}{\makecell{\textbf{VLM label: present}\\ \textbf{Human label: absent} }} \\
    \cmidrule(lr){3-4}
    \textbf{Attribute} & \textbf{Fleiss kappa} & \textbf{VLM correct} & \makecell[cl]{\textbf{VLM incorrect (categories)}} \\
    \midrule
    Age~Group & 0.693 & 64\% & animals; drawings of people; statues; dolls\\
    Gender & 0.680 & 18\% & statues; drawings of people; dolls; person visible but gender indistinguishable; helmet; toy \\
    Hair~Color & 0.650 & 22\% & drawings of people; statues; animals; doll; person visible but hair color indistinguishable \\
    Spectators & 0.641 & 70\% & a group of people is present, but no clear spectating of an event; only participants of an event\\
    \makecell[tl]{Face (Partial) \hspace{-20pt}} & 0.580 &   40\% & visible full face; animal; drawings of people; part of the body; mask; fireworks \\ 
    \bottomrule
    \end{tabular}
    \label{tab:fleiss_kappa}
    \vspace{-10pt}
\end{table}

 To analyse the disagreements between the VLMs and the human annotators, we take the top-five attributes with the highest Fleiss~kappa~\cite{fleiss1971measuring} among the VLM annotators (see Tab.~\ref{tab:fleiss_kappa}).
 The Fleiss kappa measures how well the annotators are aligned in their responses, and ranges from $-1$ (agreement worse than by chance), to $1$ (perfect agreement). Taking attributes with high Fleiss kappa allows us to avoid selecting images where the models agreed by chance.  
For each attribute, we consider the images on which models agree. This reduces the number of images per attribute to $5875$ for \textit{Age Group}, $5810$ for \textit{Gender}, $5636$ for \textit{Hair Color}, $5755$ for \textit{Spectators}, and $5407$ for \textit{Face (Partial)} attributes. We refer to these collective VLM annotations as VLM labels, and to human annotations as human labels. We compute the confusion matrices for VLM and human labels for each attribute (see Fig.~\ref{fig:confusion_matrix}) and discuss them below.

\begin{figure}
    \centering
\begin{minipage}{\linewidth}
\centering
\includegraphics[width=1\linewidth]{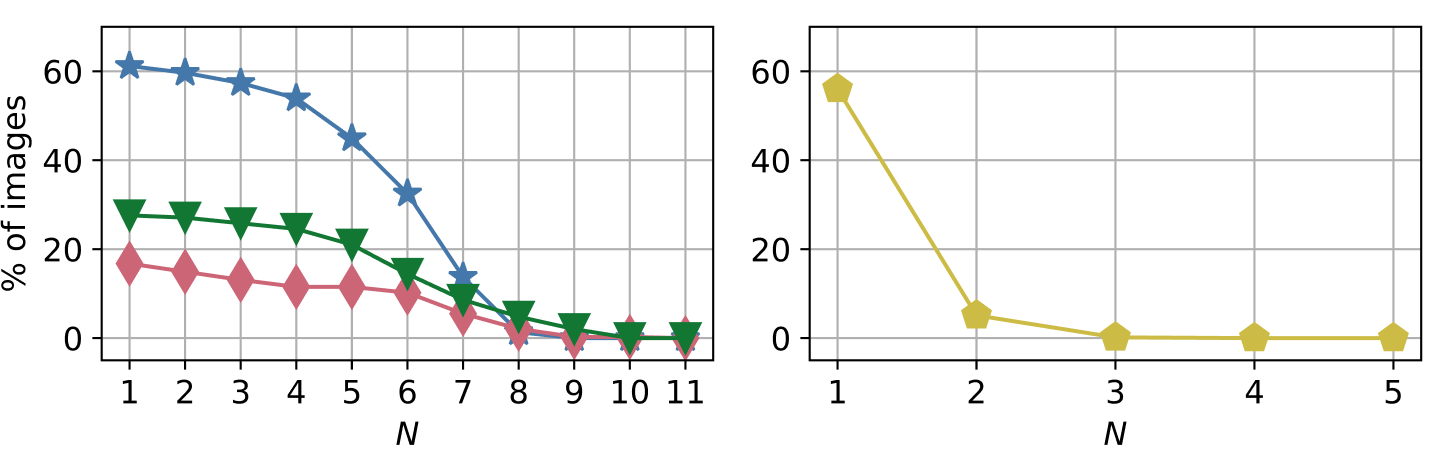}
\end{minipage}
\vspace{-12pt}
    \caption{Left: Percentage of images with at least $N$ other human-defining attributes present according to human annotators for \textit{Age~Group} (\hspace{-0.9ex}\protect\LegendCaptionLineStar{cb_blue}{cb_blue}),
     \textit{Gender}~(\hspace{-0.9ex}\protect\LegendCaptionLineDiamond{cb_pink}{cb_pink}), \textit{Hair Color}~(~\hspace{-1.7ex}\protect\LegendCaptionLineTriangleDown{cb_green}{cb_green}) attributes. Right:~Percentage of images with at least $N$ other relationship-defining attributes present according to human annotators for \textit{Spectators}~(\hspace{-0.7ex}\protect\LegendCaptionLinePentagon{cb_yellow}{cb_yellow}) attribute. The percentage is computed out of the cases when the VLM label is present, and the human label is absent (i.e., 523 for the \textit{Age Group} attribute).
    }
    \vspace{-7pt}
    \label{fig:percentage_analysis}
\end{figure}

\noindent \textbf{Age Group, Gender, and Hair Color.} We now analyse the disagreement cases for \textit{Age Group}, \textit{Gender}, and \textit{Hair Color} attributes. We investigate which other human-labelled attributes were present in the image. Out of 67 VISPR attributes, we manually select the attributes that correspond to a person being present in the image, the human-defining attributes: \textit{Age Group}, \textit{Weight Group}, \textit{Height Group}, \textit{Gender}, \textit{Eye Color}, \textit{Hair Color}, \textit{Face (Complete)}, \textit{Face (Partial)}, \textit{Semi-nudity}, \textit{Complete Nudity}, \textit{Race}, \textit{Skin Color}. If many of the human-defining attributes appear in the image annotation, it is highly likely that either \textit{Age Group}, \textit{Gender}, or \textit{Hair Color} attribute is also present, but was omitted by the human annotator. 

For \textit{Age Group}, \textit{Gender}, or \textit{Hair Color}, for the cases when the human label is absent, and the VLM label is present, we plot the percentage of images that contain at least $N$ other human-defining attributes (see Fig.~\ref{fig:percentage_analysis}, left). We notice an elbow drop at $N=6$,  which means that at least six other human-defining attributes are present in these images. We assume that the presence of the other six human-defining attributes is sufficient for \textit{Age Group}, \textit{Gender}, and \textit{Hair Color} attributes to be present in the image. For example, if \textit{Face (Partial)}, \textit{Race}, \textit{Skin Color}, \textit{Gender}, \textit{Hair Color}, and \textit{Face (Complete)} attributes are present, the \textit{Age Group} is likely to be present as well. Under this assumption, at least 167 images are missing a human label for the \textit{Age Group}: out of the images where all three VLMs annotated this attribute as present, but the human-annotated label is absent (523 images), 32\% of them include at least six other human-identifying attributes.
Similarly, 10\% of the 382 images for the \textit{Gender} attribute, and 14\% of 395 images for the \textit{Hair Color} attribute are likely to have been omitted by human annotators. 

\noindent \textbf{Spectators.} We now analyse the cases of disagreement for the \textit{Spectators} attribute. 
From the original list of attributes, we take a subset of the relationship-defining attributes: \textit{Personal Relationships}, 
\textit{Social Circle}, 
\textit{Professional Circle}, 
\textit{Competitors}, 
\textit{Spectators}, and 
\textit{Similar View}. 
We consider the cases when the VLM label is present for \textit{Spectators}, and the human label is absent. In 56\% of such cases, there exists at least one other relationship-defining attribute according to the human annotator (see Fig.~\ref{fig:percentage_analysis}, right). This means that while the models are able to detect the presence of relationship attributes, they might be in disagreement with the human annotators on the types of relationships that are present.

\noindent \textbf{Face (Partial).} To analyse the disagreement cases for this attribute, we firstly define the group of attributes that corresponds to the presence of the human face, face-defining attributes: \textit{Face (Complete)}, \textit{Face (Partial)}, and \textit{Eye Color}. 
Out of 1819 disagreement images where \textit{Face (Partial)} is present according to the VLM, and absent according to a human, 59\% of the images contain at least one other face-defining attribute annotated by a human annotator, and 46\% contain both of these attributes. While this does not guarantee that the face was partially visible, it guarantees the presence of a face.

We compare the difference in annotation for \textit{Face (Partial)} with \textit{Face (Complete)} attributes.  The annotation instructions for \textit{Face (Partial)} and \textit{Face (Complete)} are as follows: "Less than 70\% of the face is visible or there is occlusion, such as when the subject is wearing sunglasses" and "A face is completely visible. Also includes photographs of faces on identity cards, documents or billboards"~\cite{orekondy_towards_2017}.
For human annotation, both \textit{Face (Partial)} and \textit{Face (Complete)} have the same label in 75\% of the cases. A similar pattern holds for Qwen2.5-VL-7B-Instruct (77\%), Gemma-3-4b-it (84\%), and Llama-3.2-11B-Vision-Instruct (75\%). 

When either one of the attributes is present, but not both, the human annotates \textit{Face (Complete)} more often than the \textit{Face (Partial)}, which corresponds to 15\% and 10\% of the cases, respectively. 
The tendency is reversed for the Qwen2.5-VL-7B-Instruct and Llama-3.2-11B-Vision-Instruct, since when either one of the attributes is present, but not both, \textit{Face (Partial)} is annotated more often than \textit{Face (Complete)}. 
For Qwen2.5-VL-7B-Instruct, 21\% of the cases include a present label for \textit{Face (Partial)} and absent for \textit{Face (Complete)}, and only 1\% of the cases when \textit{Face (Complete)} is present but the \textit{Face (Partial)} is absent. 
Instead, for Llama-3.2-11B-Vision-Instruct, it is 19\% for \textit{Face (Partial)} being present and \textit{Face (Complete)} absent, and 5\% vice versa. 
For Gemma-3-4b-it, the model predicts only \textit{Face (Complete)}, and only \textit{Face (Partial)} 7\% and 9\% of the cases accordingly.

\noindent \textbf{Manual inspection.} For each of the five attributes, we randomly select 50 images where the VLM label is present and the human label is absent, and manually analyse them. We re-annotate the selected images with two annotators per attribute, following the VISPR definition of attributes. We define a subject in annotation instructions as a living person; therefore, we do not consider attributes related to drawings, statues, and dolls as present. The disagreements were resolved further in the discussion. As a result, we found that, for the majority of such images related to \textit{Age Group} and \textit{Spectators} attributes, VLMs correctly identify attributes that were initially overlooked by VISPR human annotators. However, VLMs erroneously detect attributes in photos of statues, drawings of people, religious illustrations, and animals.

\vspace{-11pt}

\section{Conclusion}
\label{sec:conclusion}
\vspace{-3pt}
We analysed and evaluated the ability of VLMs to recognise privacy attributes in images. We identified the attributes for which VLMs are in strong agreement when annotating, such as \textit{Age Group}, \textit{Gender}, \textit{Hair Color}, \textit{Spectators}, \textit{Face (Partial)}, and analysed disagreement between model predictions and human annotations. We found that VLMs can help to detect attributes such as \textit{Age Group}, \textit{Gender}, and \textit{Hair Color}, which are at times omitted by human annotators. Finally, we observed that VLMs often tend to annotate images with a \textit{Face (Partial)} rather than a \textit{Face (Complete)} attribute and, for the \textit{Spectators} attribute, either correctly identify it or predict related relationship attributes in the image. For these attributes, the disagreements with human annotations often stem from human annotation errors or the presence of related attributes in the image.

\bibliographystyle{IEEEbib}
\bibliography{references}

\newpage

\end{document}